\documentclass{article}

\usepackage{graphicx}
\usepackage{natbib}
\usepackage{caption}
\usepackage[utf8]{inputenc} % allow utf-8 input
\usepackage{booktabs}       % professional-quality tables
\usepackage{amsfonts}       % blackboard math symbols
\usepackage{nicefrac}       % compact symbols for 1/2, etc.
\usepackage{microtype}      % microtypography
\usepackage{xcolor}         % colors
\usepackage{amsmath}
\usepackage{amssymb}
\usepackage{siunitx}
\usepackage{calc,array} % for pandoc tables
\usepackage{hyperref}

\makeatletter
\providecommand{\tightlist}{%
  \setlength{\itemsep}{0pt}\setlength{\parskip}{0pt}}
\def\fps@figure{!ht}
\def\maxwidth{\ifdim\Gin@nat@width>\linewidth\linewidth\else\Gin@nat@width\fi}
\def\maxheight{\ifdim\Gin@nat@height>\textheight\textheight\else\Gin@nat@height\fi}
\makeatother
\setkeys{Gin}{width=\maxwidth,height=\maxheight,keepaspectratio}

\makeatletter
\newcounter{figno}
\newenvironment{fignos:no-prefix-figure-caption}{
  \caption@ifcompatibility{}{
    \let\oldthefigure\thefigure
    \let\oldtheHfigure\theHfigure
    \renewcommand{\thefigure}{figno:\thefigno}
    \renewcommand{\theHfigure}{figno:\thefigno}
    \stepcounter{figno}
    \captionsetup{labelformat=empty}
  }
}{
  \caption@ifcompatibility{}{
    \captionsetup{labelformat=default}
    \let\thefigure\oldthefigure
    \let\theHfigure\oldtheHfigure
    \addtocounter{figure}{-1}
  }
}
\makeatother

\renewcommand{\vec}[1]{\mathbf{#1}}

\title{Information Bottleneck-Based Hebbian Learning Rule Naturally Ties Working Memory and Synaptic Updates}

\author{Kyle Daruwalla$^1$ \and Mikko Lipasti$^2$}
\date{
  Electrical and Computer Engineering Department \\
  University of Wisconsin-Madison \\
  Madison, WI 53705 \\
  $^1$\texttt{daruwalla@wisc.edu}, $^2$\texttt{mikko@engr.wisc.edu}
}

\begin{document}

\maketitle

\begin{abstract}
  While deep feedforward neural networks are effective models for a wide
array of problems, back-propagation, which makes training such networks
possible, is biologically implausible. Neuroscientists are uncertain
about how the brain would propagate a precise error signal backward
through a network of neurons. Recent progress
\citep{LillicrapRandomFeedbackWeights2014, LillicrapBackpropagationBrain2020}
addresses part of this question, e.g., the weight transport problem, but
a complete solution remains intangible. In contrast, novel learning
rules
\citep{MaHSICBottleneckDeep2019, PogodinKernelizedInformationBottleneck2020}
based on the information bottleneck (IB) train each layer of a network
independently, circumventing the need to propagate errors across layers.
Instead, propagation is implicit due the layers' feedforward
connectivity. These rules take the form of a three-factor Hebbian update
--- a global error signal modulates local synaptic updates within each
layer. Unfortunately, the global signal for a given layer requires
processing multiple samples concurrently, and the brain only sees a
single sample at a time. Prior work limits the rule to an approximated
two-point update to preserve biological plausibility. Our findings show
that this restriction negatively impacts the IB estimate and
convergence. Instead, we propose a new three-factor update rule where
the global signal correctly captures information across samples via an
auxiliary reservoir network. The auxiliary network can be trained
\emph{a priori} independently of the dataset being used with the primary
network. We demonstrate comparable performance to baselines on image
classification tasks. Interestingly, unlike back-propagation-like
schemes where there is no link between learning and memory, our rule
presents a direct connection between working memory and synaptic
updates. To the best of our knowledge, this is the first rule to make
this link explicit. We explore these implications in initial experiments
examining the effect of memory capacity on learning performance. Moving
forward, this work suggests an alternate view of learning where each
layer balances memory-informed compression against task performance.
This view naturally encompasses several key aspects of neural
computation, including memory, efficiency, and locality.

\end{abstract}

\hypertarget{introduction}{%
\section{Introduction}\label{introduction}}

The success of deep learning demonstrates the usefulness of large
feedforward networks for solving a variety of tasks. Bringing the same
results to spiking neural networks is challenging, since the driving
factor behind deep learning's success --- back-propagation --- is not
considered to be biologically plausible
\citep{LillicrapBackpropagationBrain2020}. Specifically, it is unclear
how neurons might propagate a precise error signal within a
forward/backward pass framework like back-propagation. A large body of
work has been devoted to establishing plausible alternatives or
approximations for this error propagation scheme
\citep{AkroutDeepLearningWeight2019, BalduzziKickbackCutsBackprop2015, ScellierEquilibriumPropagationBridging2017, LillicrapBackpropagationBrain2020}.
While these approaches do address some of the issues with
back-propagation, implausible elements, like separate inference and
learning phases, still persist.

\begin{figure}
\hypertarget{fig:explicit-implicit-propagation}{%
\centering
\includegraphics[width=0.8\textwidth,height=\textheight]{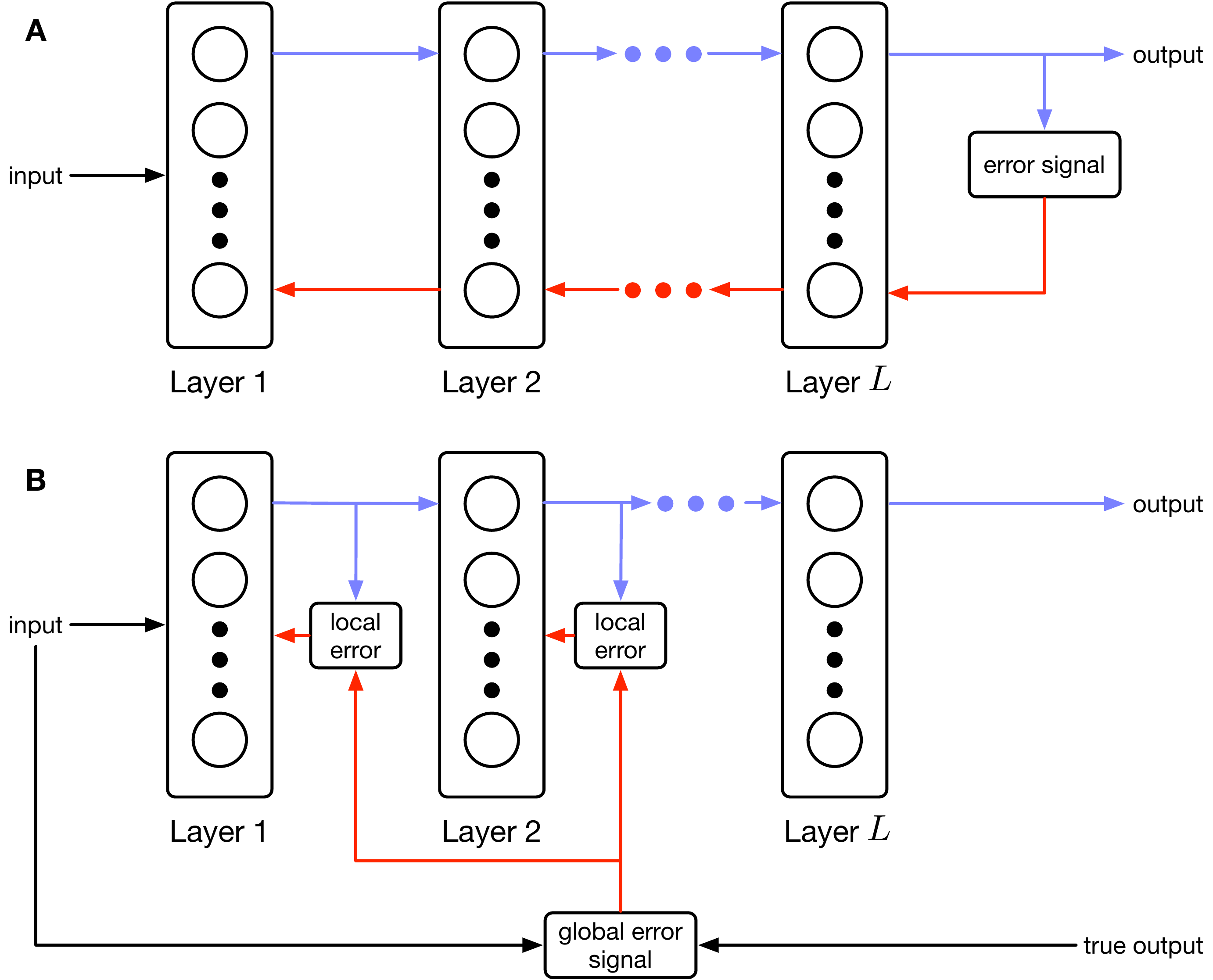}
\caption{\textbf{A.} Sequential (explicit) error propagation requires
precise information transfer \emph{backwards} between layers.
\textbf{B.} Parallel (implicit) error propagation uses only local
information in combination with a global modulating signal. Biological
rules of this form are known as three-factor learning rules
\citep{FremauxNeuromodulatedSpikeTimingDependentPlasticity2016}.}\label{fig:explicit-implicit-propagation}
}
\end{figure}

In this work, we rely on recent advances in deep learning that train
feedforward networks by balancing the information bottleneck
\citep{MaHSICBottleneckDeep2019}. Unlike back-propagation, where an
error signal computed at the end of the network is propagated to the
front (see Fig. \ref{fig:explicit-implicit-propagation}A), this method,
called the Hilbert-Schmidt Independence Criterion (HSIC) bottleneck,
applies the information bottleneck to each layer in the network
independently. Layer-wise optimization is biologically plausible as
shown in Fig. \ref{fig:explicit-implicit-propagation}B.

Our contributions include:

\begin{enumerate}
\def\labelenumi{\arabic{enumi}.}
\tightlist
\item
  We show that optimizing the HSIC bottleneck via gradient descent emits
  a three-factor learning rule
  \citep{FremauxNeuromodulatedSpikeTimingDependentPlasticity2016}
  composed of a local Hebbian component and a global layer-wise
  modulating signal.
\item
  The HSIC bottleneck depends on a batch of samples, and this is
  reflected in our update rule. Unfortunately, the brain only sees a
  single sample at a time. We show that the local component only
  requires the current sample, and that the global component can be
  accurately computed by an auxiliary reservoir network. The reservoir
  acts as a working memory, and the effective ``batch size'' corresponds
  to its capacity.
\item
  We demonstrate the empirical performance of our update rule by
  comparing it against baselines on synthetic datasets as well as MNIST
  \citep{LeCunMNISTHandwrittenDigit1998}.
\item
  To the best of our knowledge, our rule is the first to make a direct
  connection between working memory and synaptic updates. We explore
  this connection in some initial experiments on memory size and
  learning performance.
\end{enumerate}

\hypertarget{related-work}{%
\section{Related work}\label{related-work}}

Several works have presented approximations to back-propagation.
Variants of feedback alignment
\citep{AkroutDeepLearningWeight2019, LiaoHowImportantWeight2016, LillicrapRandomFeedbackWeights2014}
address the weight transport problem. Target propagation
\citep{AhmadGAITpropBiologicallyPlausible2020} and equilibrium
propagation \citep{ScellierEquilibriumPropagationBridging2017} propose
alternative mechanisms for propagating error. Yet, all these methods
require separate inference (forward) and learning (backward) phases.

\begin{figure}
\hypertarget{fig:baseline-accuracies}{%
\centering
\includegraphics[width=0.8\textwidth,height=\textheight]{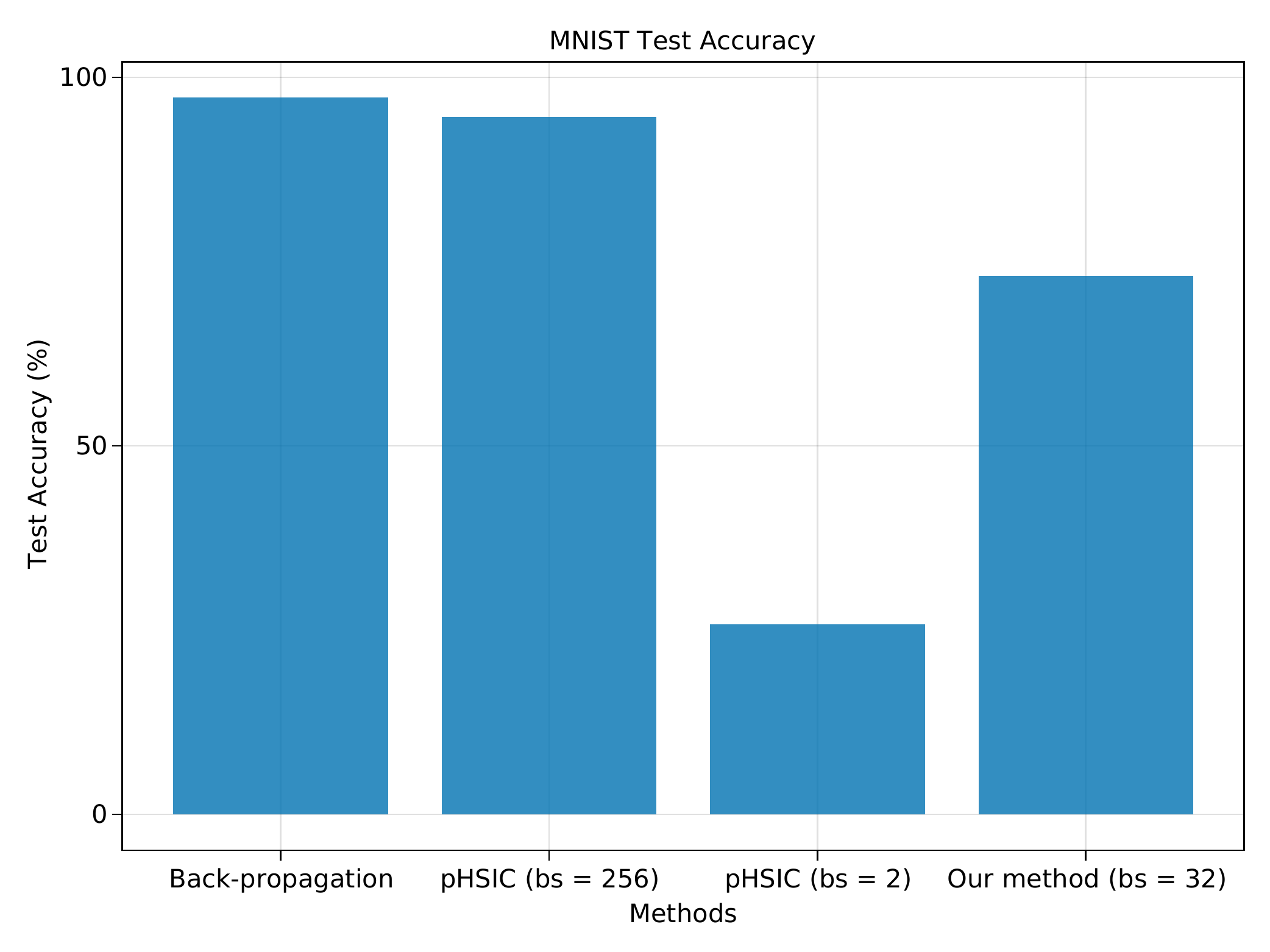}
\caption{Final test accuracies on MNIST for various learning rules.
pHSIC (plausible HSIC) is proposed in
\citet{PogodinKernelizedInformationBottleneck2020}. ``pHSIC (bs = 256)''
is the accuracy reported in that work, and ``pHSIC (bs = 2)'' is the
accuracy we obtain when applying the same rule with a reduced batch
size. The degradation in performance is our motivation for deriving an
alternate rule that effectively captures the batch size crucial to the
HSIC bottleneck.}\label{fig:baseline-accuracies}
}
\end{figure}

Layer-wise objectives
\citep{BelilovskyGreedyLayerwiseLearning2019, NoklandTrainingNeuralNetworks2019},
like the one used in this work, offer an alternative that avoids the
weight transport problem entirely. Moreover, our objective emits a
biologically plausible three-factor learning rule which can be applied
concurrently with inference.
\citet{PogodinKernelizedInformationBottleneck2020} draw similar
intuition in their work on the plausible HSIC (pHSIC) learning rule. But
in order to make experiments with the pHSIC computationally feasible,
the authors used an approximation where the network ``sees'' 256 samples
at once. In contrast, their proposed rule only receives information from
two samples --- the current one and the previous one. As shown in Fig.
\ref{fig:baseline-accuracies}, we find this limited memory capacity
reduces the effectiveness of weight updates. This motivates our work, in
which we derive an alternate rule where only the global component
depends on past samples (the local component only requires the current
pre- and post-synaptic activity). Furthermore, we show that this global
component can be computed using a reservoir network. This allows us to
achieve performance much closer to back-propagation without compromising
the biological plausibility of the rule.

\hypertarget{notation}{%
\section{Notation}\label{notation}}

First, we will introduce the notation used in the paper.

\begin{itemize}
\tightlist
\item
  Vectors are indicated in bold and lower-case (e.g.~\(\vec{x}\)).
\item
  Matrices are indicated in bold and upper-case (e.g.~\(\vec{W}\)).
\item
  Superscripts refer to different layers of a feedforward network
  (e.g.~\(\vec{z}^\ell\) is the \(\ell\)-th layer).
\item
  Subscripts refer to individual samples (e.g.~\(\vec{x}_i\) is the
  \(i\)-th sample).
\item
  Brackets refer to elements within a matrix or vector
  (e.g.~\([\vec{x}]_i\) is the \(i\)-th element of \(\vec{x}\)).
\end{itemize}

\hypertarget{the-information-bottleneck}{%
\section{The information bottleneck}\label{the-information-bottleneck}}

Given an input random variable, \(X\), an output label random variable,
\(Y\), and hidden representation, \(T\), the information bottleneck (IB)
principle is described by
\begin{equation}\min_{P_{T \mid X}} I(X; T) - \gamma I(T; Y)\label{eq:information-bottleneck}\end{equation}
where \(I(X; Y)\) is the mutual information between two random
variables. Intuitively, this expression adjusts \(T\) to achieve a
balance between information compression and output preservation.

Though computing the mutual information of two random variables requires
knowledge of their distributions, \citet{MaHSICBottleneckDeep2019}
propose using the Hilbert-Schmidt Independence Criterion (HSIC) as a
proxy for mutual information. Given a finite number of samples, \(N\), a
statistical estimator for the HSIC
\citep{GrettonMeasuringStatisticalDependence2005} is
\begin{equation}\begin{aligned}
\mathrm{HSIC}(X, Y) & = (N - 1)^{-2} \mathrm{tr}(\vec{K_X} \vec{H} \vec{K_Y} \vec{H}) \\
& = \frac{1}{(N - 1)^2} \sum_{p = 1}^N \sum_{q = 1}^N \bar{k}(\vec{x}_p, \vec{x}_q) \bar{k}(\vec{y}_q, \vec{y}_p)
\end{aligned}\label{eq:hsic}\end{equation}
\begin{equation}\begin{aligned}[]
[\vec{K_X} \vec{H}]_{pq} & = \bar{k}(\vec{x}_p, \vec{x}_q) \\
& = k(\vec{x}_p, \vec{x}_q) - \frac{1}{N} \sum_{n = 1}^N k(\vec{x}_p, \vec{x}_n)
\end{aligned}\label{eq:center-kernel-matrix}\end{equation}
\begin{equation}[\vec{K_X}]_{pq} = k(\vec{x}_p, \vec{x}_q) = \exp\left(-\frac{\|\vec{x}_p - \vec{x}_q\|^2}{\sigma^2}\right)\label{eq:uncenter-kernel-matrix}\end{equation}
where Eq. \ref{eq:center-kernel-matrix} and
\ref{eq:uncenter-kernel-matrix} define the centered and uncentered
kernel matrices, respectively.

Using these definitions \citet{MaHSICBottleneckDeep2019} define the HSIC
objective --- a loss function for training feedforward neural networks
by balancing the IB at each layer. Consider a feedforward neural network
with \(L\) layers where the output of layer \(\ell\) is
\[\vec{z}^\ell = f(\vec{W}^\ell \vec{z}^{\ell - 1})\] We train the
network to minimize \begin{equation}\begin{aligned}
\mathcal{L}_{\mathrm{HSIC}}(X, Y, Z^\ell) & = \mathrm{HSIC}(X, Z^\ell) - \gamma \mathrm{HSIC}(Y, Z^\ell) \\
& \qquad \forall \ell \in \{1, \ldots, L\}
\end{aligned}\label{eq:hsic-objective}\end{equation} where
\(Z = \{Z^\ell\}_{\ell = 1}^L\) are the output distributions at each
hidden layer. Note that there is a separate objective for each layer. As
a result, there is no explicit error propagation across layers, and the
error propagation is implicit due to forward connectivity as shown in
Fig. \ref{fig:explicit-implicit-propagation}.

\hypertarget{deriving-a-biologically-plausible-rule-for-the-hsic-bottleneck}{%
\section{Deriving a biologically-plausible rule for the HSIC
bottleneck}\label{deriving-a-biologically-plausible-rule-for-the-hsic-bottleneck}}

In this work, we seek to optimize Eq. \ref{eq:hsic-objective} for a
feedforward network of leaky integrate-and-fire (LIF) neurons. Given the
membrane potentials, \(\vec{u}^\ell\), for layer \(\ell\), the dynamics
are governed by \begin{equation}\begin{aligned}
    \tau_{\mathrm{ff}} \frac{\mathrm{d}\vec{u}^\ell}{\mathrm{d}t} &= -\vec{u} + \vec{W^\ell} \vec{z}^{\ell - 1} \\
    \vec{z}^\ell &= \tanh(\vec{u}^\ell) + \zeta
\end{aligned}\label{eq:lif-dynamics}\end{equation} where
\(\vec{z}^\ell\) is the output activity (firing rate) and
\(\zeta \sim \mathrm{Unif}(-0.05, 0.05)\) emulates the firing rate
noise.

We optimize \(\mathcal{L}_{\mathrm{HSIC}}\) by taking the gradient with
respect to \(\vec{W}^\ell\) and applying gradient descent. Doing this,
we obtain the following update rule: \begin{equation}\begin{aligned}
    \Delta [\vec{W}^\ell]_{ij} &\propto \sum_{p, q = 0}^{-(N - 1)}
        \left[\bar{k}(\vec{x}_p, \vec{x}_q) - \gamma \bar{k}(\vec{y}_p, \vec{y}_q)\right] \bar{\alpha}_{ij}(\vec{z}_p^\ell, \vec{z}_q^\ell) \\
    \bar{\alpha}_{ij}(\vec{z}_p^\ell, \vec{z}_q^\ell) &=
            \alpha_{ij}(\vec{z}_p^\ell, \vec{z}_q^\ell) - \frac{1}{N} \sum_{n = 0}^{-(N - 1)} \alpha_{ij}(\vec{z}_p^\ell, \vec{z}_n^\ell) \\
    \alpha_{ij}(\vec{z}_p^\ell, \vec{z}_q^\ell) &=
        -\frac{k(\vec{z}_p^\ell, \vec{z}_q^\ell)}{\sigma^2}
        ([\vec{z}_p^\ell]_i - [\vec{z}_q^\ell]_i) \\
        &\qquad \left(\frac{\partial \vec{z}_p^\ell}{\partial [\vec{W}^\ell]_{ij}}
            - \frac{\partial \vec{z}_q^\ell}{\partial [\vec{W}^\ell]_{ij}}\right)
\end{aligned}\label{eq:weight-update-implausible}\end{equation} where
\(\alpha_{ij}\) is a Hebbian-like term between pre-synaptic neuron \(j\)
and post-synaptic neuron \(i\). Note that the indices, \(p\) and \(q\),
are from zero to \(-(N - 1)\) to indicate samples at previous time steps
(i.e.~\(\vec{z}_0^\ell\) corresponds to the layer output now,
\(\vec{z}_{-1}^\ell\) corresponds to the layer output from the
previously seen sample, etc.). We call \(N\), the batch size in the deep
learning, the \emph{effective batch size} in our work. A full derivation
of Eq. \ref{eq:weight-update-implausible} can be found in the appendix.
This rule is similar to the basic rule in
\citet{PogodinKernelizedInformationBottleneck2020}, except that they
replace \(\bar{k}(\vec{x}_p, \vec{x}_q)\) with
\(\bar{k}(\vec{z}_p, \vec{z}_q)\) and do not use a centered kernel
matrix.

Without modifications, Eq. \ref{eq:weight-update-implausible} is not
biologically plausible. \(\alpha_{ij}(\vec{z}_p^\ell, \vec{z}_q^\ell)\)
cannot be called Hebbian when \(p\) and \(q\) are not equal to zero,
since it depends on non-local information from the past. We solve this
by making a simplifying approximation. We assume that
\(\partial \vec{z}_p^\ell / \partial [\vec{W}^\ell]_{ij} = 0\) when
\(p \neq 0\). In other words, the weights at the current time step do
not affect past outputs. With this assumption, we find that
\(\alpha_{ij}(\vec{z}_p^\ell, \vec{z}_q^\ell) \neq 0\) if and only if
\((p = 0, q \neq 0)\) or \((p \neq 0, q = 0)\). This gives us our final
update rule: \begin{equation}\begin{aligned}
    \Delta [\vec{W}^\ell]_{ij} &\propto \beta_{ij} \xi_i \\
    \beta_{ij} &= (1 - ([\vec{z}_0^\ell]_i)^2) [\vec{z}_0^{\ell - 1}]_j \\
    \xi_{i} &= \sum_{p = 0}^{-(N - 1)} \left[\bar{k}(\vec{x}_0, \vec{x}_p) - \gamma \bar{k}(\vec{y}_0, \vec{y}_p)\right] \bar{\alpha}_i(\vec{z}_p^\ell) \\
    \bar{\alpha}_i(\vec{z}_p^\ell) &= \alpha_i(\vec{z}_p^\ell) - \frac{1}{N} \sum_{n = 0}^{-(N - 1)} \alpha_i(\vec{z}_n^\ell) \\
    \alpha_i(\vec{z}_p^\ell) &= -\frac{2 k(\vec{z}_0^\ell, \vec{z}_p^\ell)}{\sigma^2} (\vec{z}_0^\ell - \vec{z}_p^\ell)
\end{aligned}\label{eq:weight-update}\end{equation} Details for deriving
Eq. \ref{eq:weight-update} from Eq. \ref{eq:weight-update-implausible}
are found in the appendix. Note that \(\beta_{ij}\) is now a Hebbian
term that only depends on the current pre- and post-synaptic activity.
\(\xi_i\) is a modulating term that adjusts the synaptic update
layer-wise. This establishes a three-factor learning rule for Eq.
\ref{eq:hsic-objective}. Yet, it is still not biologically plausible,
since the global error, \(\xi_i\), requires buffering layer activity
over many samples. We discuss how to overcome this problem next.

\hypertarget{computing-the-modulating-signal-with-a-reservoir-network}{%
\subsection{Computing the modulating signal with a reservoir
network}\label{computing-the-modulating-signal-with-a-reservoir-network}}

In order to compute \(\xi_i\) in Eq. \ref{eq:weight-update}
biologically, we require a neural circuit capable of storing information
for future use. Recurrent networks can provide such functionality, and
\citet{HoerzerEmergenceComplexComputational2014} demonstrate how a
reservoir network can be trained to compute complex signals using a
binary teaching signal.

For each layer, we construct an auxiliary network of LIF neurons whose
dynamics are governed by \begin{equation}\begin{aligned}
    \tau_r \frac{\mathrm{d}\vec{u}_r}{\mathrm{d}t} &= -\vec{u}_r + \lambda \vec{W}_r \vec{r} + \vec{W}_i \vec{r}_i + \vec{W}_{\mathrm{fb}} \vec{r}_o \\
    \vec{r} &= \tanh(\vec{u}_r) + \zeta_r \\
    \vec{r}_o &= \vec{W}_o \vec{u}_r + \zeta_o
\end{aligned}\label{eq:reservoir-dynamics}\end{equation} where
\(\vec{u}_r\) are the recurrent neuron membrane potentials,
\(\vec{r}_i\) is the input signal activity, and \(\vec{r}_o\) is the
readout activity. \(\zeta_r\) is the recurrent firing rate noise and
\(\zeta_o\) is the exploratory noise for the readout activity.
\(\lambda\) is a hyper-parameter that controls the chaos level of the
recurrent population.

As stated in \citet{HoerzerEmergenceComplexComputational2014}, we train
the auxiliary network using a three-factor Hebbian learning rule shown
below, \begin{equation}\begin{aligned}
    \Delta \vec{W}_o &\propto M (\vec{r}_o - \bar{\vec{r}}_o) \vec{r}^\top \\
    M &= \begin{cases}
        1 & P > \bar{P} \\
        0 & P \leq \bar{P}
    \end{cases} \\
    P &= -\|\vec{r}_o - \xi\|_2^2
\end{aligned}\label{eq:rm-hebb}\end{equation} where \(\xi\) is the
\emph{true} global error signal in Eq. \ref{eq:weight-update}, \(P\) is
the negative mean-squared error of the current network output vs.~the
true signal, \(M\) is a binary teaching signal derived from \(P\), and
\(\bar{P}\), \(\bar{\vec{r}}_o\) are low-pass filtered versions of
\(P\), \(\vec{r}o\), respectively. The low-pass filter averages the
signals over a moving window (based on the same function as
\citet{HoerzerEmergenceComplexComputational2014}). Intuitively, this
rule uses Hebbian updates to train the readout weights, but it gates the
strength of the updates based on whether the network performance is
improving or not.

\begin{figure}
\hypertarget{fig:network-architecture}{%
\centering
\includegraphics[width=0.85\textwidth,height=\textheight]{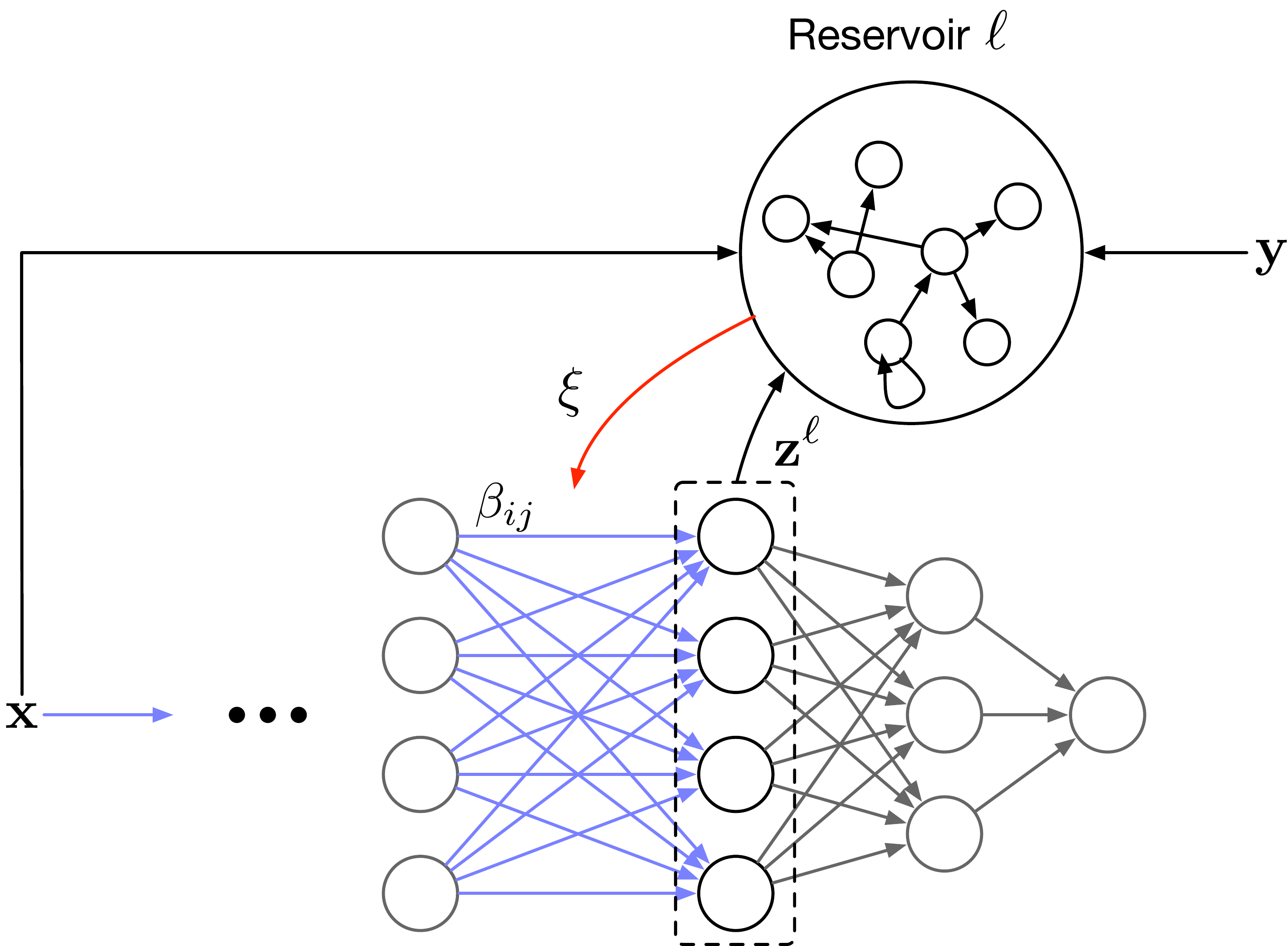}
\caption{The overall network architecture. Each layer has a
corresponding auxiliary reservoir network. The synaptic update,
\(\beta_{ij}\) in Eq. \ref{eq:weight-update}, is modulated by a
layer-wise error signal, \(\xi\), that is the readout from the
reservoir.}\label{fig:network-architecture}
}
\end{figure}

Fig. \ref{fig:network-architecture} illustrates the full design of the
proposed learning scheme. The reservoir serves as a working memory where
the capacity of the memory determines the effective batch size. To the
best of our knowledge, our rule is the first to modulate the Hebbian
updates of a synapse based on past information stored in a working
memory. Furthermore, having a controllable effective batch size means we
can study the effect of memory capacity on the learning convergence. In
particular, we can compare performance against the \(N = 2\) case which
matches the biologically plausible variant of the prior work
\citep{PogodinKernelizedInformationBottleneck2020}.

\hypertarget{experiments}{%
\section{Experiments}\label{experiments}}

We tested our design through a series of experiments on the reservoir
individually and the full network on various synthetic and standard
datasets. The code to reproduce each experiment is available at
\url{https://github.com/darsnack/biological-hsic} along with instructions.
Experiments were performed on a single node machine with a AMD Ryzen
Threadripper 1950X 16-Core Processor and Nvidia Titan Xp GPU.

All data is normalized to \([0, 1]\) to represent rate-encoded signals.
Simulations use a time step of \SI{1}{\ms}.

\begin{figure}
\hypertarget{fig:mnist-class-predictions}{%
\centering
\includegraphics{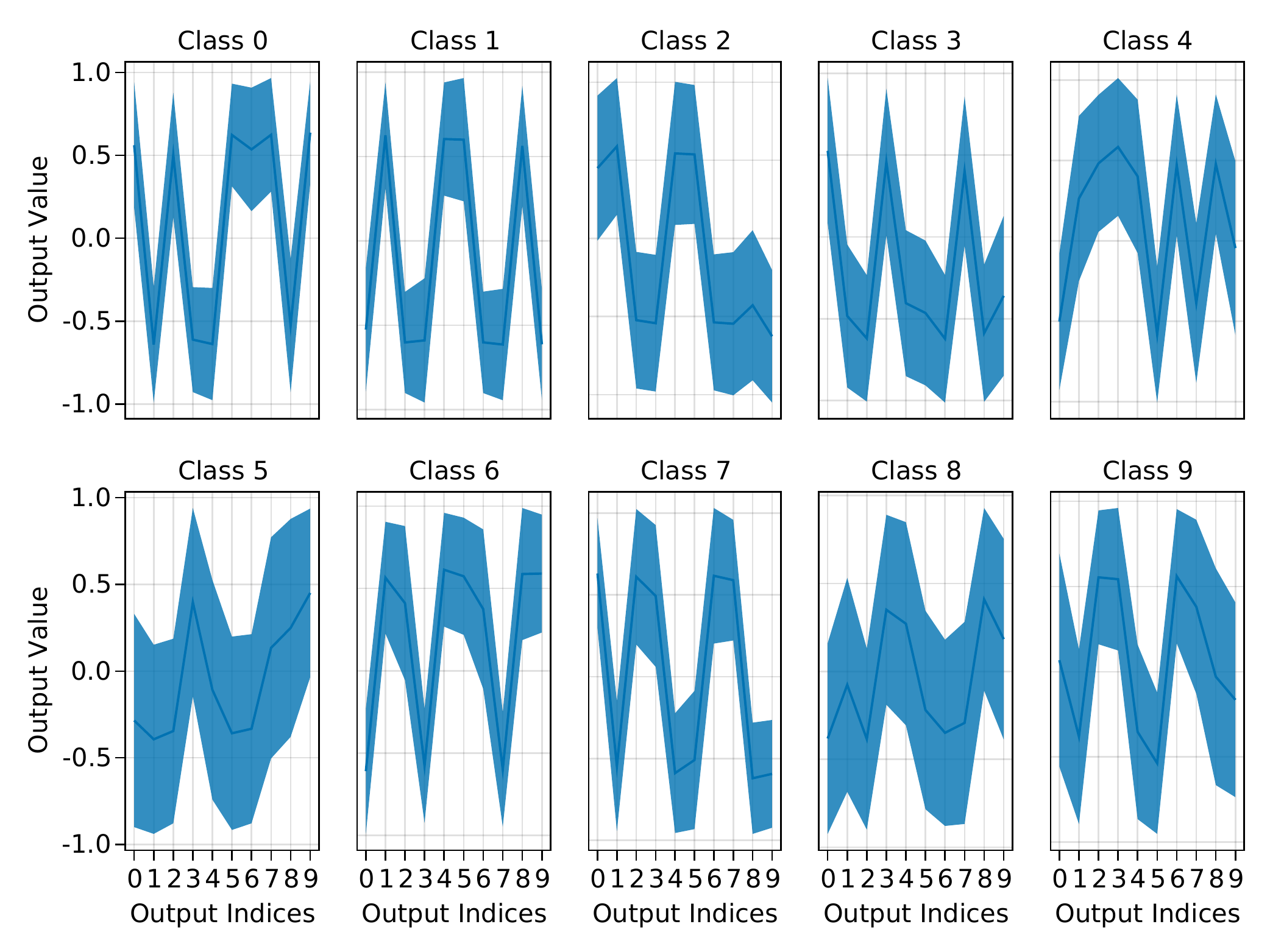}
\caption{The average predicted output across all test samples on MNIST
separated by true label (the shaded region indicates the standard
deviation across test samples). Each subplot corresponds to a single
true label, and the solid line is the mean output for the final
HSIC-trained layer. As expected, the output is not one-hot, but the
network does tend to learn a unique binary ``code word'' for each
class.}\label{fig:mnist-class-predictions}
}
\end{figure}

The learned output of the final layer via Eq. \ref{eq:hsic-objective} is
not necessarily one-hot like the true labels. This is because the HSIC
objective only attempts to match the predicted output distribution and
the true output distribution based on similarity between representations
(this behavior is explained later in Fig.
\ref{fig:mnist-class-predictions}). Like
\citet{MaHSICBottleneckDeep2019}, we use a linear readout layer trained
for 1000 epochs with gradient descent to map between the HSIC-learned
output and the label encoding. This is only required for our
experiments. Biological circuitry would not require a specific label
encoding.

\hypertarget{reservoir-experiments}{%
\subsection{Reservoir experiments}\label{reservoir-experiments}}

First, we verify the ability for the reservoir to reproduce the true
signal \(\xi\). We use a recurrent population of 2000 neurons with
\(\tau_r =\) \SI{50}{\ms} and \(\lambda = 1.2\) (based on
recommendations in \citet{HoerzerEmergenceComplexComputational2014}). A
hundred random input signals, \(X \in \mathbb{R}^{100 \times 100}\),
\(Y \in \mathbb{R}^{1 \times 100}\), and
\(Z \in \mathbb{R}^{10 \times 100}\), are drawn from
\(\mathrm{Unif}(0, 1)\). We train the reservoir for \SI{500}{\s} to
learn \(\xi\) based on \(X, Y, Z\), then stop the weight updates and
test it for \SI{100}{\s}. The following learning rate decay schedule is
used

\[\eta(t) = \frac{\eta_0}{1 + \frac{t}{\tau_{\text{decay}}}}\]

A complete list of parameters is in Table
\ref{tbl:reservoir-experiment-parameters}.

\begin{table}[ht]

\centering

\begin{tabular}{lcr} \toprule

\textbf{Parameter Name} & \textbf{Symbol} & \textbf{Value} \\ \midrule

LPF Time Constant & \(\tau_{\mathrm{lpf}}\) & \SI{5}{\ms} \\
Sample Time Constant & \(\Delta t_{\text{sample}}\) & \SI{50}{\ms} \\
Hidden Firing Noise & \(\zeta_r\) & \num{5e-6} \\
Readout Firing Noise & \(\zeta_o\) & \num{1e-2} \\
Effective Batch Size & \(N\) & \num{6} \\
HSIC Balance Parameter & \(\gamma\) & \num{2} \\
Learning Rate & \(\eta_0\) & \num{1e-4} \\
Learning Rate Decay & \(\tau_{\text{decay}}\) & \SI{20}{\s} \\

\bottomrule

\end{tabular}

\caption{Reservoir experiment parameters.
\label{tbl:reservoir-experiment-parameters}}

\end{table}

\begin{figure}
\hypertarget{fig:reservoir-experiment}{%
\centering
\includegraphics{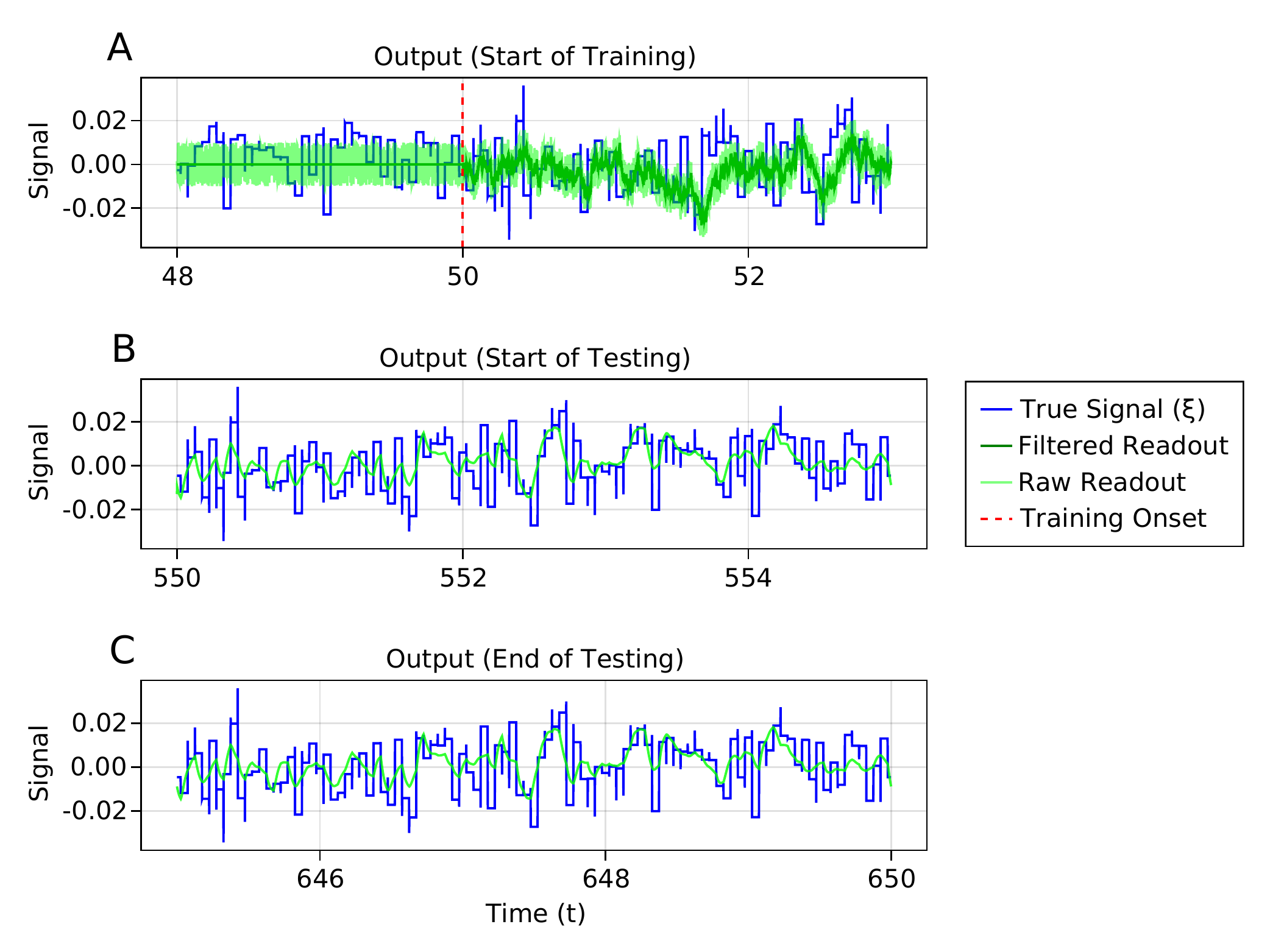}
\caption{The reservoir output when learning \(\xi\) in Eq.
\ref{eq:weight-update}. \textbf{A.} The output at the right before and
after the start of training. When the training begins, the readout
quickly begins to match the true signal. \textbf{B.} The output at the
start of testing. Even without the feedback from the learning rule, the
output matches the true signal. \textbf{C.} The output at the end of
testing (\SI{100}{\s} later) continues to maintain the correct
behavior.}\label{fig:reservoir-experiment}
}
\end{figure}

The results can be seen in Fig. \ref{fig:reservoir-experiment} which
shows the reservoir output for the first element of the output signal.
Prior to training, the output is pure noise, but it begins to match the
true signal quickly after training begins. Even when the training is
stopped, the reservoir continues to produce the correct output for the
full testing period.

\hypertarget{small-dataset-experiments}{%
\subsection{Small dataset experiments}\label{small-dataset-experiments}}

In order to demonstrate that a reservoir readout can effectively
modulate learning for a complete network, we perform a series of
small-scale experiments. These experiments use simple synthetic datasets
through which we can gain an intuitive understanding of the learning
behavior. In addition to the parameters listed in Table
\ref{tbl:reservoir-experiment-parameters}, Table
\ref{tbl:small-experiment-parameters} contains shared parameters for all
small-scale experiments.

\begin{table}[ht]

\centering

\begin{tabular}{lcr} \toprule

\textbf{Parameter Name} & \textbf{Symbol} & \textbf{Value} \\ \midrule

Network LIF Time Constant & \(\tau_{\mathrm{ff}}\) & \SI{5}{\ms} \\
Reservoir Chaos Level & \(\lambda\) & \num{1.7} \\
Hidden Firing Noise & \(\zeta_r\) & \num{5e-2} \\
Readout Firing Noise & \(\zeta_o\) & \num{2.5e-1} \\
\# of Hidden Neurons & \(N_{\text{hidden}}\) & \num{1000} \\
Effective Batch Size & \(N\) & \num{10} \\
Network Learning Rate & \(\eta_0\) & \num{5e-3} \\
Network Learning Rate Decay & \(\tau_{\text{decay}}\) & \SI{50}{\s} \\

\bottomrule

\end{tabular}

\caption{Small-scale experiment parameters.
\label{tbl:small-experiment-parameters}}

\end{table}

\begin{figure}
\hypertarget{fig:linear-experiment}{%
\centering
\includegraphics{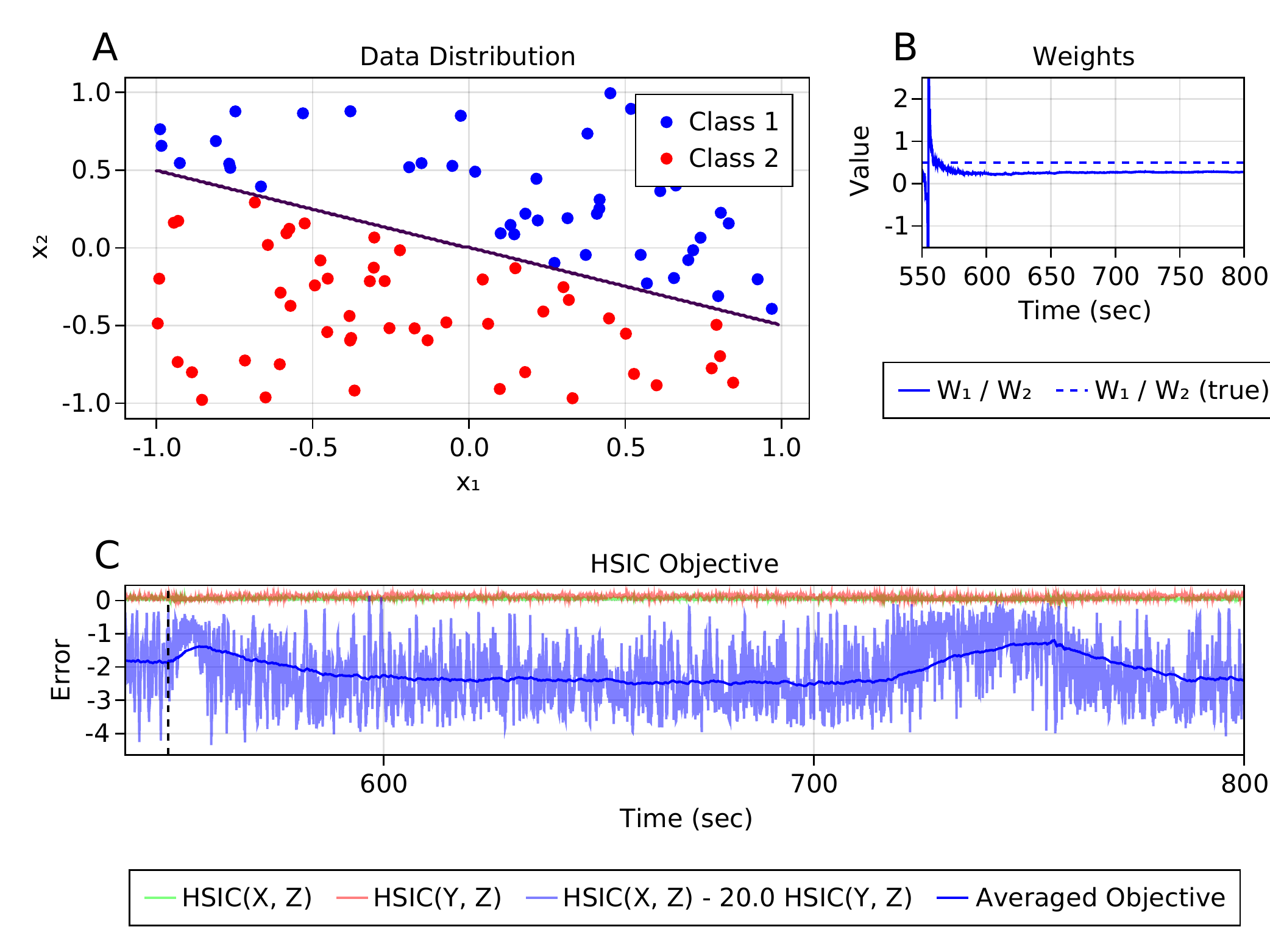}
\caption{The results from training a single layer perceptron on a linear
binary classification task. \textbf{A.} An example sample set from the
synthetic distribution. \textbf{B.} The ratio of the two weights in the
model converges the true ratio. \textbf{C.} The HSIC objective converges
due to the learning rule. The final test accuracy is
94\%.}\label{fig:linear-experiment}
}
\end{figure}

First, we test a single layer perceptron on a linear binary
classification task. We separated 100 uniformly sampled points from
\([-1, 1] \times [-1, 1]\) using a line as shown in Fig.
\ref{fig:linear-experiment}A. We pre-train the reservoir network for
\SI{550}{\s} with \(\gamma = 5\), then we perform 50 epochs of training
on the dataset. The network reached a final classification accuracy of
94\%. Furthermore, the ratio of the learned weights converges towards
the ratio of the true coefficients of the linear boundary.

\begin{figure}
\hypertarget{fig:tanh-experiment}{%
\centering
\includegraphics{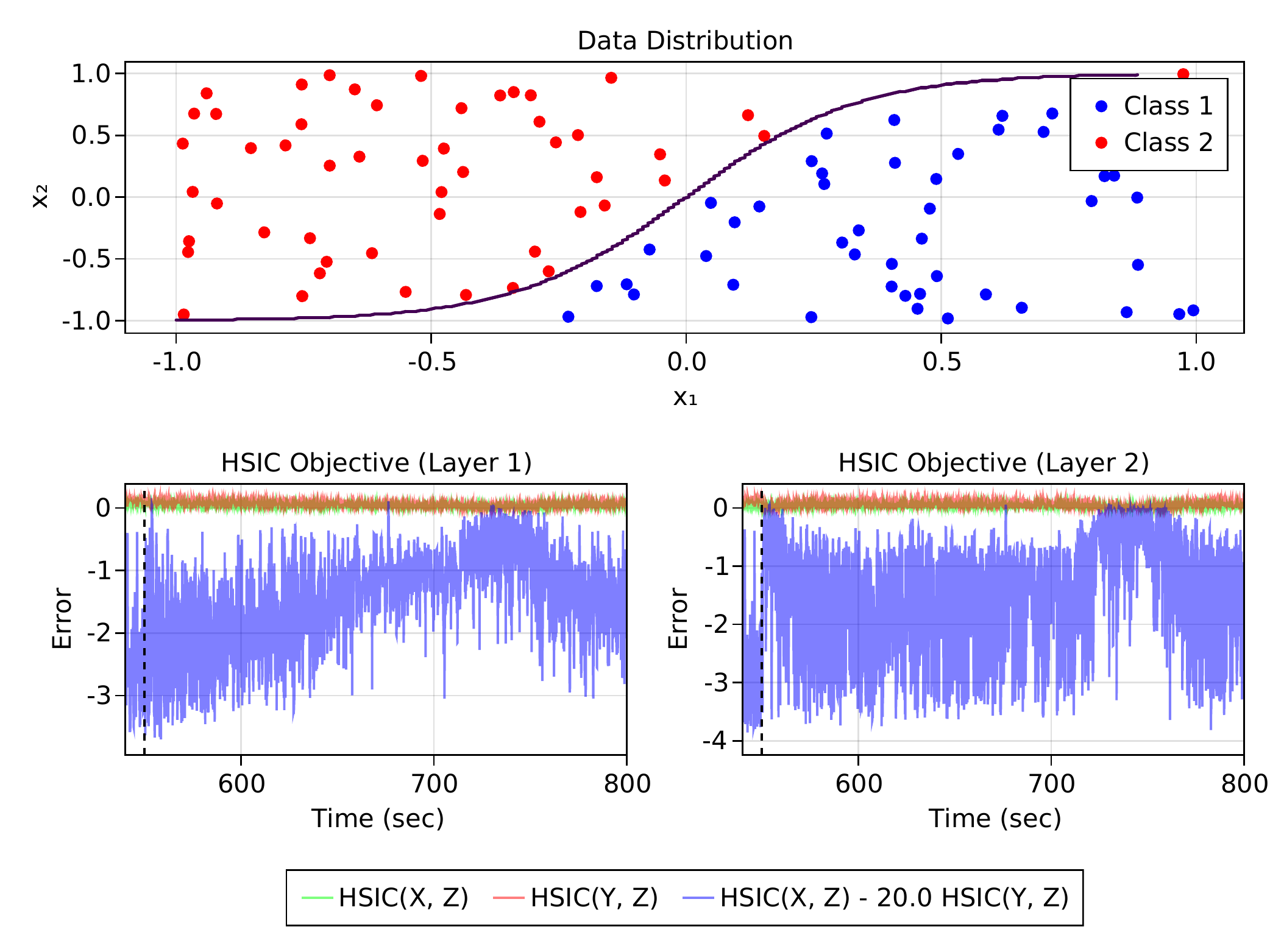}
\caption{The results from training a two-layer MLP on a synthetic binary
classification dataset. \textbf{A.} An example sample set from the
synthetic distribution. \textbf{B.} The HSIC objective (Eq.
\ref{eq:hsic-objective}) for each layer. The plain objective value is
very noisy due to the noise in the neuron firing rate. Despite this
noise, the network converges and reaches 98\%
accuracy.}\label{fig:tanh-experiment}
}
\end{figure}

Next, we test a multilayer perceptron (MLP) network with an auxiliary
reservoir in place for each layer. A synthetic 2D binary classification
dataset is generated by sampling 100 random points in
\([-1, 1] \times [-1, 1]\) uniformly, then separating the points by the
decision boundary \(x_2 = \tanh(3x_1)\). An example generated dataset is
shown in Fig. \ref{fig:tanh-experiment}A. The MLP architecture is two
layers of size two and one, respectively. We warm-up and pre-train the
reservoir network for \SI{550}{\s} with \(\gamma = 20\), then we run 50
epochs of training with our rule. On testing, the network achieves 98\%
accuracy.

In both small experiments, there is a noticeable oscillation in the
objective value during training. We attribute this instability to the
output of the reservoir falling out of phase with the true signal before
returning in-phase. This behavior was noted in
\citet{HoerzerEmergenceComplexComputational2014}, and it is periodic. In
our experiments, we disable weight updates to the reservoir to
demonstrate the ability for the full framework to learn even with an
imperfect modulating signal. In practice, the reservoir can be
continuously updated, which will eliminate the phase shift
\citep{HoerzerEmergenceComplexComputational2014}.

\hypertarget{large-dataset-experiments}{%
\subsection{Large dataset experiments}\label{large-dataset-experiments}}

Finally, we test our rule on MNIST with an MLP against a
back-propagation baseline. The architecture used for all learning
frameworks is a
\(784 \Rightarrow 64 \Rightarrow 32 \Rightarrow 10 \Rightarrow 10\)
network. The back-propagation baseline uses ReLU activation and
artificial neurons like those typically used in deep learning. Our
method uses LIF neurons as described in Sec.
\ref{deriving-a-biologically-plausible-rule-for-the-hsic-bottleneck}.

Since our learning rule is designed to process single samples at a time,
we must iterate and train on MNIST a single sample at a time. This
becomes computationally infeasible due to the large dataset size. In
order to make the experiments tractable, we do not simulate the
reservoir neurons. Having already established their ability to capture
the global signal, \(\xi\), correctly, we compute the signal
analytically. Furthermore, we obtain a subset of MNIST by sampling 50\%
of the points in the training data via stratified sampling. We retain
the complete test data for reporting final accuracy. The
back-propagation network is trained with a batch size of 32, and our
rule uses an effective batch size of 32. The networks are all trained
for 75 epochs, and each sample is presented for \SI{20}{\ms}. The HSIC
balance parameter, \(\gamma\), is set to 50, 100, 200, and 500, for each
layer, respectively. We use a momentum optimizer (with momentum,
\(\rho = 0.9\)) purely to speed up convergence and make the experiment
tractable. The learning rate starts at \num{5e-2} for 20 epochs, drops
to \num{5e-3} for 30 epochs, and finally \num{1e-3} for the remaining
training time. The remaining parameters match the previous experiments.

\begin{figure}
\hypertarget{fig:mnist-accuracy}{%
\centering
\includegraphics[width=0.8\textwidth,height=\textheight]{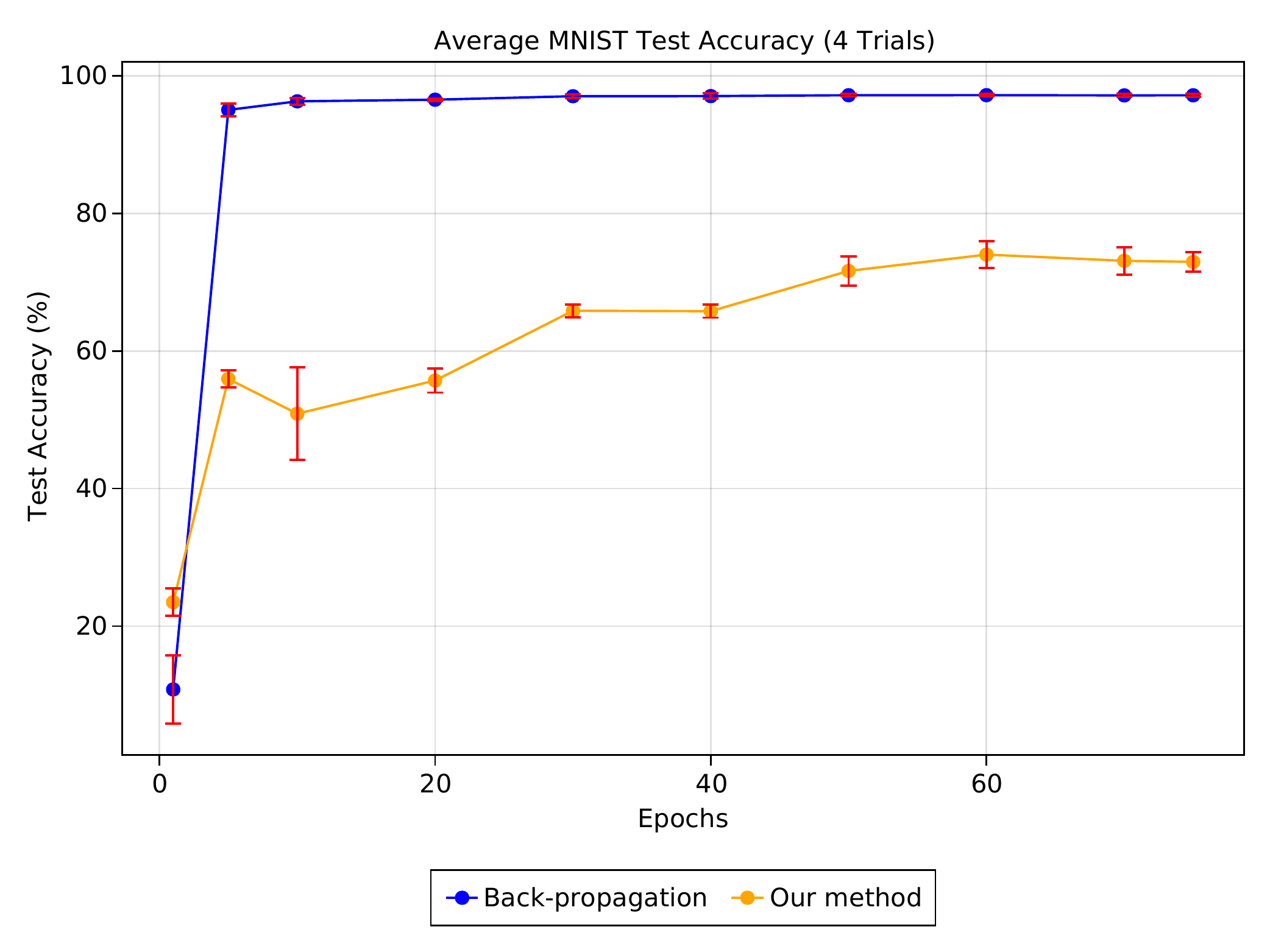}
\caption{Average test accuracy over four independent trials on MNIST for
back-propagation and our method. The red error bars indicate the
standard deviation across trials. Our method improves in accuracy as
training proceeds, but convergence quickly slows down. We expect the gap
in performance to close with more epochs but running for many epochs was
computationally intractable for this work.}\label{fig:mnist-accuracy}
}
\end{figure}

We performed four trials to account for random seeds and show the
average test accuracy across trials in Fig. \ref{fig:mnist-accuracy}.
Despite failing to match the same accuracy as back-propagation, our
method is able to improve and learn the dataset. After 75 epochs, our
objective value continued to decrease but at a diminished rate. We
expect that the gap between our method and back-propagation would be
smaller with more epochs of training; however, our goal is not to
establish a state-of-the-art method for training. Instead, we are
demonstrating a biologically-plausible rule that can scale up to larger
datasets.

In addition to examining the test accuracy, we show the average
predicted output per class from one the trials in Fig.
\ref{fig:mnist-class-predictions}. We see that the predictions do not
match a one-hot vector, but each label is assigned a unique binary
``code word.''

Note that we repeat these experiments with pHSIC (batch size = 2) and
LIF neurons to produce the result in Fig. \ref{fig:baseline-accuracies}.

\begin{figure}
\hypertarget{fig:mnist-subset-accuracy}{%
\centering
\includegraphics[width=0.8\textwidth,height=\textheight]{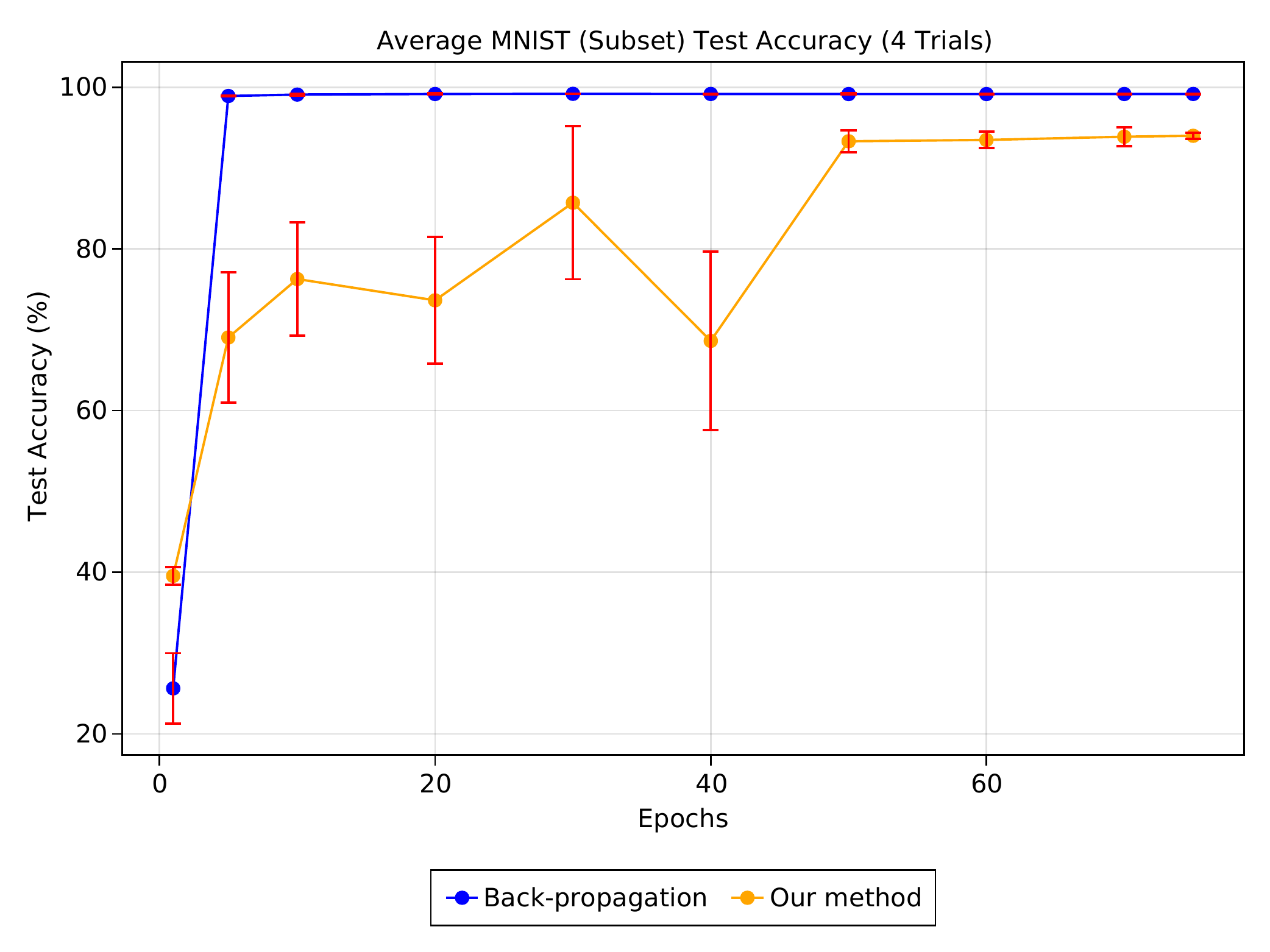}
\caption{The average test accuracy across 4 trials on a subset of MNIST
(only digits 0, 1, 2, and 4). Our method (94\% accuracy) achieves
comparable performance to
back-propagation.}\label{fig:mnist-subset-accuracy}
}
\end{figure}

In order to demonstrate that our rule can scale up to larger datasets
with sufficient iterations, we create a subset of MNIST by only
selecting samples for the digits 0, 1, 2, and 4. We repeat our
experiment on this subset of data. The results are shown in Fig.
\ref{fig:mnist-subset-accuracy}. With a slightly smaller, but still
complex dataset, our method achieves comparable performance to
back-propagation.

\hypertarget{effects-of-memory-capacity}{%
\subsection{Effects of memory
capacity}\label{effects-of-memory-capacity}}

\begin{figure}
\hypertarget{fig:memory-sweep}{%
\centering
\includegraphics[width=0.8\textwidth,height=\textheight]{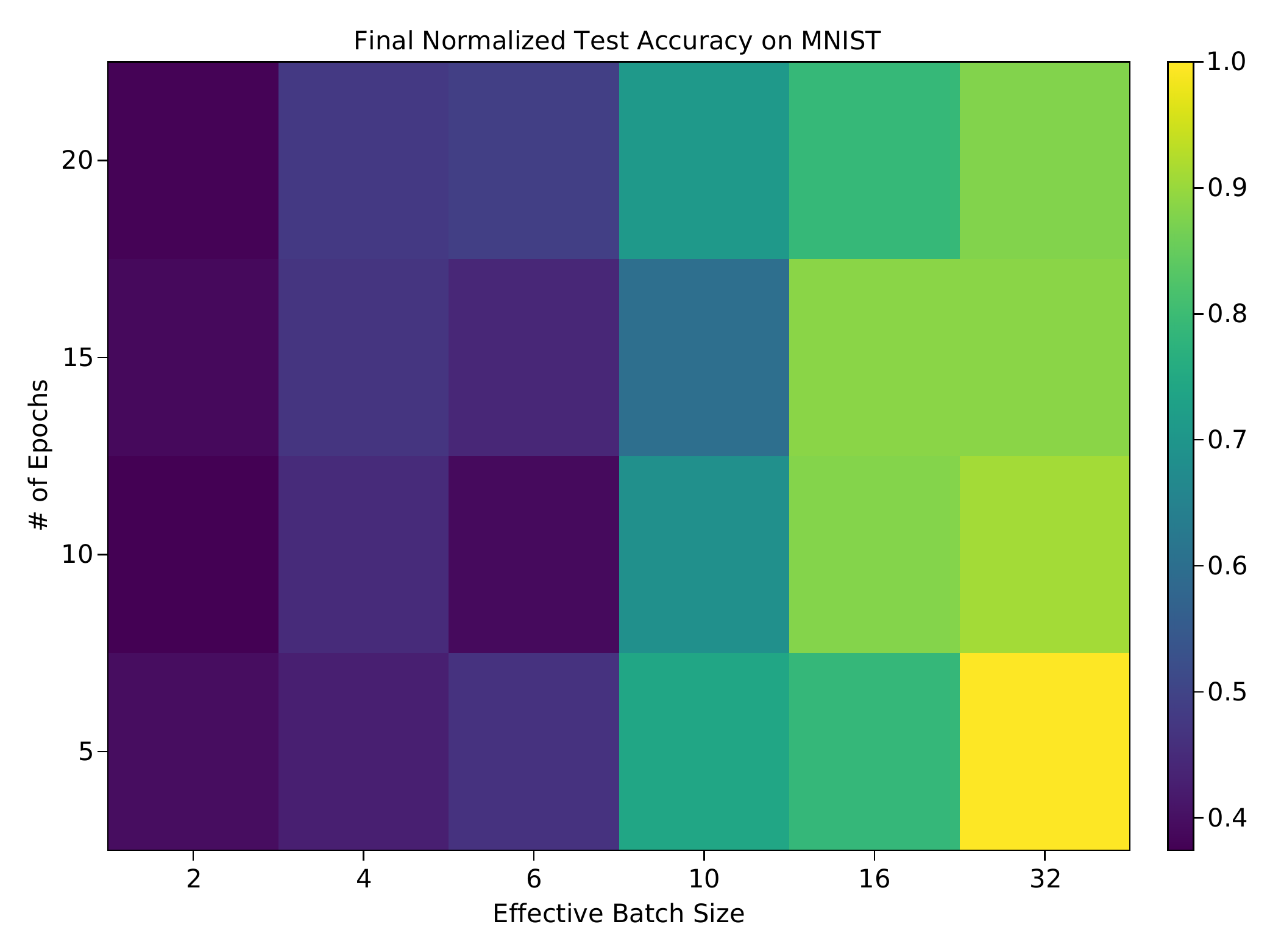}
\caption{The final normalized test accuracy on MNIST for varying number
of epochs and effective batch sizes. We see that the accuracy is
degraded for small batch sizes independent of the number of
epochs.}\label{fig:memory-sweep}
}
\end{figure}

One of the novel features of our rule is the ability to control the
memory capacity of the update. To explore this parameter, we repeat the
same MNIST experiments as before for various effective batch sizes and
number of epochs of training. The results are shown in Fig.
\ref{fig:memory-sweep}. The normalized final test accuracy is much lower
for small batch sizes independent of the number of training epochs. This
is because for a given random variable, \(X \in \mathbb{R}^n\), the
kernel matrix, \(\vec{K}_X\), is the basis for an estimate of how
samples of \(X\) are distributed in \(\mathbb{R}^n\). When the effective
batch size is small, this estimate is poor and provides an erroneous
signal for modulating the weight updates. In particular, note that
\(\alpha_i(\vec{z}_p^\ell)\) in Eq. \ref{eq:weight-update} has an
anti-Hebbian behavior --- it tends to drive \(\vec{z}_0^\ell\) and
\(\vec{z}_p^\ell\) (the latent representation for layer \(\ell\)) apart.
This signal is flipped whenever \(\bar{k}(\vec{y}_0, \vec{y}_p)\) is
large. In other words, if \(\vec{y}_0\) is more similar to \(\vec{y}_p\)
than it is to other samples, the latent representations are driven
towards each other. When the batch size is small, \(\vec{K}_Y\) rarely
contains samples of the same label, so the rule tends to over-drive the
latent representations apart.

\hypertarget{discussion}{%
\section{Discussion}\label{discussion}}

In this work, we proposed a three-factor learning rule for training
feedforward networks based on the information bottleneck principle. The
rule is biologically plausible, and we are able to scale up to
reasonable performance on MNIST without compromise. We do this by
factoring our weight update into a local component and global component.
The local component depends only on the current synaptic activity, so it
can be implemented via Hebbian learning. In contrast to prior work, our
global component uses information across many samples seen over time. We
show that this content can be stored in an auxiliary reservoir network,
and the readout of the reservoir can be used to modulate the local
weight updates. To the best of our knowledge, this is the first
biological learning rule to tightly couple the synaptic updates with a
working memory capacity. We verified the efficacy of our rule on
synthetic datasets and MNIST, and we explored the effect of the size of
the working memory capacity on the learning performance.

Even though our rule does perform reasonably well, there is room for
improvement. The rule performs best when it is able to distinguish
between different high dimensional inputs. The resolution at which it
separates inputs is controlled by the parameter, \(\sigma\), in the
kernel function (Eq. \ref{eq:uncenter-kernel-matrix}). The use of a
fixed \(\sigma\) is partly responsible for the slow down in convergence
in Fig. \ref{fig:mnist-accuracy}. In \citet{MaHSICBottleneckDeep2019},
the authors propose using multiple networks trained with the different
values of \(\sigma\) and averaging the output across networks. This
allows the overall network to separate the data at different
resolutions. Future work can consider a population of networks with
varying \(\sigma\) to achieve the same effect. Alternatively, the lower
hierarchies of the visual cortex have built-in pre-processing at varying
spatial resolutions (e.g.~Gabor filters). We could consider adding a
series of filters before the network and pass the concatenated output to
the network. Addressing the resolution issue will be important for
improving the speed and scalability of the learning method.

Additionally, our rule is strongly supervised. While the mechanism for
synaptic updates is biologically plausible, the overall learning
paradigm is not. Note that the purpose of the label information in the
global signal is to indicate whether the output for the current sample
should be the same or different from previous samples. In other words,
it might be possible to replace the term
\(\bar{k}(\vec{y}_0, \vec{y}_p)\) in Eq. \ref{eq:weight-update} with a
binary teaching signal. This would allow the rule to operate under weak
supervision.

Most importantly, while our rule is certainly biologically plausible, it
remains to be seen if it is an accurate model for circuitry in the
brain. Since rules based on the information bottleneck are relatively
new, the corresponding experimental evidence must still be obtained.
Yet, we note that our auxiliary reservoir serves a similar role to the
``blackboard'' circuit proposed in
\citet{MumfordComputationalArchitectureNeocortex1991}. This circuit,
present in the thalamus, receives projected connections from the visual
cortex, similar to how each layer projects its output onto the
reservoir. Furthermore, Mumford suggests that this circuit acts as a
temporal buffer and sends signals that capture information over longer
timescales back to the cortex like our reservoir.

So, while it is uncertain whether our exact rule and memory circuit are
present in biology, we suggest that an in-depth exploration of
memory-modulated learning rules is necessary. We hope this work will be
an important step in that direction.

\appendix

\hypertarget{derivation-of-learning-rule}{%
\section{Derivation of learning
rule}\label{derivation-of-learning-rule}}

Below is a complete derivation of our update rule. First, we find the
derivative of \(\mathcal{L}_{\mathrm{HSIC}}\) (Eq. 5) with respect to
the layer weight, \(\vec{W}^\ell\). Then, we show the assumptions
necessary to make this update biologically plausible.

First, we re-index the summation in Eq. 2 such that samples
\(\{1, 2, \ldots, N\}\) are now samples \(\{0, -1, \ldots, -(N - 1)\}\)
over time.

\begin{equation}\mathrm{HSIC}(X, Y) = \frac{1}{(N - 1)^2} \sum_{p, q = 0}^{-(N - 1)} \bar{k}(\vec{x}_p, \vec{x}_q) \bar{k}(\vec{y}_q, \vec{y}_p)\label{eq:hsic-reindexed}\end{equation}

In other words, the first sample in the batch is the current sample, and
the last sample in the batch is the one presented \(N - 1\) samples ago.
Now, taking the derivative of Eq. 5 with respect to
\([\vec{W}]_{ij}^\ell\):

\begin{equation}\begin{aligned}
& \:\quad \frac{\partial \mathcal{L}_{\mathrm{HSIC}}(X, Y, Z^\ell)}{\partial [\vec{W}]_{ij}^\ell} \\
&= \frac{\partial}{\partial [\vec{W}]_{ij}^\ell} \Big[\mathrm{HSIC}(X, Z^\ell) - \lambda \mathrm{HSIC}(Y, Z^\ell)\Big] \\
&= \frac{\partial}{\partial [\vec{W}]_{ij}^\ell} \left[\frac{1}{(N - 1)^2} \sum_{p, q = 0}^{-(N - 1)} \bar{k}(\vec{z}_p^\ell, \vec{z}_q^\ell) \bar{k}(\vec{x}_q, \vec{x}_p)\right. \\
&\qquad \qquad \qquad \qquad \qquad \qquad - \gamma \bar{k}(\vec{z}_p^\ell, \vec{z}_q^\ell) \bar{k}(\vec{y}_q, \vec{y}_p)\Bigg] \\
&= \frac{1}{(N - 1)^2} \sum_{p, q = 0}^{-(N - 1)} \Big[\bar{k}(\vec{x}_q, \vec{x}_p) - \gamma \bar{k}(\vec{y}_q, \vec{y}_p)\Big] \frac{\partial \bar{k}(\vec{z}_p^\ell, \vec{z}_q^\ell)}{\partial [\vec{W}]_{ij}^\ell}
\end{aligned}\label{eq:derivative}\end{equation}

Focusing on the derivative of
\(\bar{k}(\vec{z}_p^\ell, \vec{z}_q^\ell)\),

\[\begin{aligned}
\frac{\partial \bar{k}(\vec{z}_p^\ell, \vec{z}_q^\ell)}{\partial [\vec{W}]_{ij}^\ell}
&= \frac{\partial}{\partial [\vec{W}]_{ij}^\ell} \left[k(\vec{z}_p^\ell, \vec{z}_q^\ell) - \frac{1}{N} \sum_{n = 0}^{-(N - 1)} k(\vec{z}_p^\ell, \vec{z}_n^\ell)\right] \\
&=  \frac{\partial k(\vec{z}_p^\ell, \vec{z}_q^\ell)}{\partial [\vec{W}]_{ij}^\ell} - \frac{1}{N} \sum_{n = 0}^{-(N - 1)} \frac{\partial k(\vec{z}_p^\ell, \vec{z}_n^\ell)}{\partial [\vec{W}]_{ij}^\ell}
\end{aligned}\]

And finally,

\[\begin{aligned}
\frac{\partial k(\vec{z}_p^\ell, \vec{z}_q^\ell)}{\partial [\vec{W}]_{ij}^\ell}
&= \frac{\partial}{\partial [\vec{W}]_{ij}^\ell} \left[\exp\left(-\frac{\|\vec{z}_p^\ell - \vec{z}_q^\ell\|^2}{2 \sigma^2}\right)\right] \\
&= -\frac{k(\vec{z}_p^\ell, \vec{z}_q^\ell)}{2 \sigma^2} \left(\frac{\partial \|\vec{z}_p^\ell - \vec{z}_q^\ell\|^2}{\partial [\vec{W}]_{ij}^\ell}\right) \\
&= -\frac{k(\vec{z}_p^\ell, \vec{z}_q^\ell) ([\vec{z}_p]_i^\ell - [\vec{z}_q]_i^\ell)}{\sigma^2} \\
&\quad \left(\frac{\partial \vec{z}_p^\ell}{\partial [\vec{W}]_{ij}^\ell} - \frac{\partial \vec{z}_q^\ell}{\partial [\vec{W}]_{ij}^\ell}\right)
\end{aligned}\]

This gives us the update rule form in Eq. 7. The term
\(\frac{\partial \vec{z}_p^\ell}{\partial [\vec{W}]_{ij}^\ell} - \frac{\partial \vec{z}_q^\ell}{\partial [\vec{W}]_{ij}^\ell}\)
is the local component, while the rest can be globally computer per
layer. But this rule is not biologically plausible, since the local
indices \(p\) and \(q\) sample neuron firing rates from past time steps.

We make the assumption that
\(\frac{\partial \vec{z}_p^\ell}{\partial [\vec{W}]{ij}^\ell} = 0\) when
\(p \neq 0\) (similarly when \(q \neq 0\)). In other words, the past
output does not depend on the current weights. With this assumption, we
note that

\[\frac{\partial \vec{z}_p^\ell}{\partial [\vec{W}]_{ij}^\ell} \begin{cases}
(1 - [\vec{z}_0^\ell]_i)^2 [\vec{z}_0^{\ell - 1}]_j & p = 0 \\
0 & p \neq 0
\end{cases}\]

Additionally, when both \((p = 0, q = 0)\),

\[\frac{\partial \vec{z}_p^\ell}{\partial [\vec{W}]_{ij}^\ell} - \frac{\partial \vec{z}_q^\ell}{\partial [\vec{W}]_{ij}^\ell} = 0\]

Thus, we find that each term in the summation in Eq. \ref{eq:derivative}
is non-zero only when \((p \neq 0, q = 0)\) and \((p = 0, q \neq 0)\).
Furthermore, due to the symmetry of the terms, we can reduce the
double-summation into twice times a single-summation. This gives us our
final form

\[\begin{aligned}
    \Delta [\vec{W}^\ell]_{ij} &\propto \beta_{ij} \xi_i \\
    \beta_{ij} &= (1 - ([\vec{z}_0^\ell]_i)^2) [\vec{z}_0^{\ell - 1}]_j \\
    \xi_{i} &= \sum_{p = 0}^{-(N - 1)} \left[\bar{k}(\vec{x}_0, \vec{x}_p) - \gamma \bar{k}(\vec{y}_0, \vec{y}_p)\right] \bar{\alpha}_i(\vec{z}_p^\ell) \\
    \bar{\alpha}_i(\vec{z}_p^\ell) &= \alpha_i(\vec{z}_p^\ell) - \frac{1}{N} \sum_{n = 0}^{-(N - 1)} \alpha_i(\vec{z}_n^\ell) \\
    \alpha_i(\vec{z}_p^\ell) &= -\frac{2 k(\vec{z}_0^\ell, \vec{z}_p^\ell)}{\sigma^2} (\vec{z}_0^\ell - \vec{z}_p^\ell)
\end{aligned}\]

where
\(\beta_{ij} = \frac{\partial \vec{z}_0^\ell}{\partial [\vec{W}]_{ij}^\ell}\)
is factored out of the summation since it no longer depends on \(p\).
This results in a local component, \(\beta_{ij}\), that can be computed
using only the current pre- and post-synaptic activity. The global
component, \(\xi\), requires memory, so we use an auxiliary network to
compute it.

\hypertarget{reservoir-network-details}{%
\section{Reservoir network details}\label{reservoir-network-details}}

Here we explain the low-pass filtering (LPF) details not covered by the
behavior described in Sec. 5, Eq. 9. As part of the learning rule in Eq.
10, the readout and error signals are low-pass filtered according to

\begin{equation}\bar{f}(t) = \left(1 - \frac{\Delta t}{\tau_{\mathrm{lpf}}}\right) \bar{f}(t - \Delta t) + \frac{\Delta t}{\tau_{\mathrm{lpf}}} f(t)\label{eq:lpf}\end{equation}

where \(\Delta t\) is the simulation time step. This is the same
filtering technique presented in
\citet{HoerzerEmergenceComplexComputational2014}.

\bibliography{./ref.bib}

\end{document}